\documentclass[11pt]{article}
\usepackage{authblk}
\usepackage{eacl2017}
\usepackage{times}
\usepackage{url}
\usepackage{latexsym}
\usepackage{amssymb}
\usepackage{gensymb}
\usepackage{hhline}

\usepackage{times}
\usepackage{latexsym}
\usepackage[fleqn]{amsmath}
\usepackage[pdftex]{graphicx}
\usepackage{color}

\DeclareMathOperator*{\argmax}{arg\,max}

\eaclfinalcopy 
\def\eaclpaperid{70} 

\title{Neural Machine Translation by Minimising the Bayes-risk with Respect to Syntactic Translation Lattices}

\author[1]{\bf Felix Stahlberg}
\author[1,2]{\bf Adri\`a de Gispert}
\author[1,2]{\bf Eva Hasler}
\author[1,2]{\bf Bill Byrne}
\affil[1]{Department of Engineering, University of Cambridge, UK}
\affil[2]{SDL Research, Cambridge, UK}
\affil[  ]{\tt	\{fs439,ad465,ech57,wjb31\}@cam.ac.uk}
\affil[ ]{\tt  \{agispert,ehasler,bbyrne\}@sdl.com}

\date{}

\begin{document}
\maketitle
\begin{abstract}
  We present a novel scheme to combine neural machine translation (NMT) with traditional statistical machine translation (SMT). Our approach borrows ideas from linearised lattice minimum Bayes-risk decoding for SMT. The NMT score is combined with the Bayes-risk of the translation according the SMT lattice. This makes our approach much more flexible than $n$-best list or lattice rescoring as the neural decoder is not restricted to the SMT search space. We show an efficient and simple way to integrate risk estimation into the NMT decoder which is suitable for word-level as well as subword-unit-level NMT. We test our method on English-German and Japanese-English and report significant gains over lattice rescoring on several data sets for both single and ensembled NMT. The MBR decoder produces entirely new hypotheses far beyond simply rescoring the SMT search space or fixing UNKs in the NMT output.
\end{abstract}

\section{Introduction}

Lattice minimum Bayes-risk (LMBR) decoding has been applied successfully to translation lattices in traditional SMT to improve translation performance of a single system~\cite{mbr-smt,lmbr-tromble,lmbr-blackwood}. However, minimum Bayes-risk (MBR) decoding is also a very powerful framework for combining diverse systems~\cite{mbr-combi,mbr-combi-adria}. Therefore, we study combining traditional SMT and NMT in a hybrid decoding scheme based on MBR. We argue that MBR-based methods in their present form are not well-suited for NMT because of the following reasons:

\begin{itemize}
\item Previous approaches work well with rich lattices and diverse hypotheses. However, NMT decoding usually relies on beam search with a limited beam and thus produces very narrow lattices~\cite{decoding-diverse1,decoding-diverse2}.
\item NMT decoding is computationally expensive. Therefore, it is difficult to collect the statistics needed for risk calculation for NMT.
\item The Bayes-risk in SMT is usually defined for complete translations. Therefore, the risk computation needs to be restructured in order to integrate it in an NMT decoder which builds up hypotheses from left to right.
\end{itemize}

To address these challenges, we use a special loss function which is computationally tractable as it avoids using NMT scores for risk calculation. We show how to reformulate the original LMBR decision rule for using it in a word-based NMT decoder which is not restricted to an $n$-best list or a lattice. Our hybrid system outperforms lattice rescoring on multiple data sets for English-German and Japanese-English. We report similar gains from applying our method to subword-unit-based NMT rather than word-based NMT.

\section{Combining NMT and SMT by Minimising the Lattice Bayes-risk}

We propose to collect statistics for MBR from a potentially large translation lattice generated with SMT, and use the $n$-gram posteriors as additional score in NMT decoding. The LMBR decision rule used by~Tromble et al.\ \shortcite{lmbr-tromble} has the form

\begin{equation}
\hat{\mathbf{y}}=\argmax_{\mathbf{y}\in\mathcal{Y}_h} \Big( \underbrace{\Theta_0 |\mathbf{y}|+\sum_{\mathbf{u}\in \mathcal{N}} \Theta_{|\mathbf{u}|}\#_\mathbf{u}(\mathbf{y})P(\mathbf{u}|\mathcal{Y}_e)}_{:=E(\mathbf{y})} \Big)
\label{eq:lmbr}
\end{equation}
where $\mathcal{Y}_h$ is the {\em hypothesis space} of possible translations, $\mathcal{Y}_e$ is the {\em evidence space} for computing the Bayes-risk, and $\mathcal{N}$ is the set of all $n$-grams in $\mathcal{Y}_e$ (typically, $n=1\dots 4$). In this work, our evidence space $\mathcal{Y}_e$ is a translation lattice generated with SMT. The function $\#_\mathbf{u}(\mathbf{y})$ counts how often $n$-gram $\mathbf{u}$ occurs in translation $\mathbf{y}$. $P(\mathbf{u}|\mathcal{Y}_e)$ denotes the path posterior probability of $\mathbf{u}$ in $\mathcal{Y}_e$. Our aim is to integrate these $n$-gram posteriors into the NMT decoder since they correlate well with the presence of $n$-grams in reference translations~\cite{ngram-confidence}. We call the quantity to be maximised the {\em evidence} $E(\mathbf{y})$ which corresponds to the (negative) Bayes-risk which is normally minimised in MBR decoding. We emphasize that this risk can be computed for any translation hypothesis and not only those produced by the SMT system.

NMT assigns a probability to a translation $\mathbf{y}=y_1^T$ of source sentence $\mathbf{x}$ via a left-to-right factorisation based on the chain rule:

\begin{equation}
P_{NMT}(y_1^T|\mathbf{x})=\prod_{t=1}^T \underbrace{P_{NMT}(y_t|y_1^{t-1},\mathbf{x})}_{=g(y_{t-1},s_t,c_t)}
\label{eq:nmt}
\end{equation}
where $g(\cdot)$ is a neural network using the hidden state of the decoder network $s_t$ and the context vector $c_t$ which encodes relevant parts of the source sentence~\cite{bahdanau}.\footnote{We refer to Bahdanau et al.~\shortcite{bahdanau} for a full discussion of attention-based NMT.} $P_{NMT}(\cdot)$ can also represent an ensemble of NMT systems in which case the scores of the individual systems are multiplied together to form a single distribution. Applying the LMBR decision rule in Eq.~\ref{eq:lmbr} directly to NMT would involve computing $P_{NMT}(\mathbf{y}|\mathbf{x})$ for all translations in the evidence space. In case of LMBR this is equivalent to rescoring the entire translation lattice exhaustively with NMT. However, this is not feasible even for small lattices because the evaluation of $g(\cdot)$ is computationally very expensive. Therefore, we propose to calculate the Bayes-risk over SMT translation lattices using only pure SMT scores, and bias the NMT decoder towards low-risk hypotheses. Our final combined decision rule is

\begin{equation}
\hat{\mathbf{y}}=\argmax_\mathbf{y} \Big(E(\mathbf{y}) + \lambda \log P_{NMT}(\mathbf{y}|\mathbf{x}) \Big).
\label{eq:combi}
\end{equation}
If $\mathbf{y}$ contains a word not in the NMT vocabulary, the NMT model provides a score and updates its decoder state as for an unknown word. We note that $E(\mathbf{y})$ can be computed even if $\mathbf{y}$ is not in the SMT lattice. Therefore, Eq.~\ref{eq:combi} can be used to generate translations outside the SMT search space. We further note that Eq.~\ref{eq:combi} can be derived as an instance of LMBR under a modified loss function.

\section{Left-to-right Decoding}
\label{sec:decoding}

Beam search is often used for NMT because the factorisation in Eq.~\ref{eq:nmt} allows to build up hypotheses from left to right. In contrast, our definition of the {\em evidence} in Eq.~\ref{eq:lmbr} contains a sum over the (unordered) set of all $n$-grams. However, we can rewrite our objective function in Eq.~\ref{eq:combi} in a way which makes it easy to use with beam search.

\begin{equation}
\begin{aligned}
 & E(\mathbf{y}) + \lambda \log P_{NMT}(\mathbf{y}|\mathbf{x})\\
=& \Theta_0 |\mathbf{y}|+\sum_{\mathbf{u}\in \mathcal{N}} \Theta_{|\mathbf{u}|}\#_\mathbf{u}(\mathbf{y})P(\mathbf{u}|\mathcal{Y}_e) \\
 &+ \lambda \sum_{t=1}^T \log P_{NMT}(y_t|y_1^{t-1},\mathbf{x}) \\
=& \sum_{t=1}^T \Big( \Theta_0 + \sum_{n=1}^4 \Theta_n P(y_{t-n}^t|\mathcal{Y}_e) \\
 &+ \lambda \log P_{NMT}(y_t|y_1^{t-1},\mathbf{x}) \Big)
\end{aligned}
\label{eq:decoding}
\end{equation}
for $n$-grams up to order 4. This form lends itself naturally to beam search: at each time step, we add to the previous partial hypothesis score both the log-likelihood of the last token according the NMT model, and the partial MBR gains from the current $n$-gram history. Note that this is similar to applying (the exponentiated scores of) an interpolated language model based on $n$-gram posteriors extracted from the SMT lattice. In the remainder of this paper, we will refer to decoding according Eq.~\ref{eq:decoding} as {\em MBR-based}  NMT.

\begin{table*}
\small
\centering

\begin{tabular}{|ll|r|r|r|}
\hline
 \textbf{Setup} && \textbf{news-test2014} & \textbf{news-test2015} & \textbf{news-test2016} \\ \hline
 \multicolumn{2}{|l|}{SMT baseline~\cite[HiFST]{hifst-grammar}}  & 18.9 & 21.2 & 26.0 \\  \hline
Single NMT (word) & Pure NMT & 17.7 & 19.6 & 23.1 \\
 & 100-best rescoring & 20.6 & 22.5 & 27.5 \\
 & Lattice rescoring & 21.6 & 23.8 & {\bf 29.6} \\
 & {\bf This work} & {\bf 22.0} & {\bf 24.6} & 29.5 \\ \hline
5-Ensemble NMT (word) & Pure NMT & 19.4 & 21.8 & 25.4 \\
 & 100-best rescoring & 21.0 & 23.3 & 28.6 \\
 & Lattice rescoring & 22.1 & 24.2 & 30.2 \\
  & {\bf This work} & {\bf 22.8} & {\bf 25.4} & {\bf 30.8} \\ \hline
  Single NMT (BPE) & Pure NMT & 19.6 & 21.9 & 24.6 \\
 & Lattice rescoring &  21.5 &  24.0 &  {\bf 29.6} \\
 & {\bf This work} & {\bf 21.7} & {\bf  24.1} &  28.6 \\ 
 \hline
3-Ensemble NMT (BPE) & Pure NMT &  21.0 &  23.4 &  27.0 \\
 & Lattice rescoring &  21.7 &  24.2 &  {\bf 30.0} \\
   & {\bf This work} & {\bf 22.3} & {\bf 24.9} &  29.2 \\
  \hline
\end{tabular}
\caption{\label{tab:results-ende} English-German lower-cased BLEU scores calculated with \texttt{mteval-v13a.pl}.\footnotemark}
\end{table*}

\section{Efficient $n$-gram Posterior Calculation}

The risk computation in our approach is based on posterior probabilities $P(\mathbf{u}|\mathcal{Y}_e)$ for $n$-grams $\mathbf{u}$ which we extract from the SMT translation lattice $\mathcal{Y}_e$. $P(\mathbf{u}|\mathcal{Y}_e)$ is defined as the sum of the path probabilities $P_{SMT}(\cdot)$ of paths in $\mathcal{Y}_e$  containing $\mathbf{u}$~\cite[Eq.~2]{lmbr-blackwood}:

\begin{equation}
P(\mathbf{u}|\mathcal{Y}_e) = \sum_{\mathbf{y}\in\{\mathbf{y}\in\mathcal{Y}_e: \#_\mathbf{u}(\mathbf{y})>0\}} P_{SMT}(\mathbf{y}|\mathbf{x}).
\end{equation}

We use the framework of Blackwood et al.~\shortcite{lmbr-blackwood} based on $n$-gram mapping and path counting transducers to efficiently pre-compute all non-zero values of $P(\mathbf{u}|\mathcal{Y}_e)$. Complete enumeration of all $n$-grams in a lattice is usually feasible even for very large lattices~\cite{lmbr-blackwood}. Additionally, for all these $n$-grams $\mathbf{u}$, we smooth $P(\mathbf{u}|\mathcal{Y}_e)$ by mixing it with the uniform distribution to flatten the distribution and increase the offset to $n$-grams which are not in the lattice.

\section{Subword-unit-based NMT}

Character-based or subword-unit-based NMT \cite{subword-huffman,subword-bpe,char-noseg,char-hybrid,char-conv,char-cmu,gnmt} does not use isolated words as modelling units but applies a finer grained tokenization scheme. One of the main motivation for these approaches is to overcome the limited vocabulary in word-based NMT. We consider our hybrid system as an alternative way to fix NMT OOVs. However, our method can also be used with subword-unit-based NMT. In this work, we use byte pair encodings~\cite[BPE]{subword-bpe} to test combining word-based SMT with subword-unit-based NMT via both lattice rescoring and MBR. First, we construct a finite state transducer (FST) which maps word sequences to BPE sequences. Then, we convert the word-based SMT lattices to BPE-based lattices by composing them with the mapping transducer and projecting the output tape using standard OpenFST operations~\cite{openfst}. The converted lattices are used for extracting $n$-gram posteriors as described in the previous sections. Note that even though the $n$-grams are on the BPE level, their posteriors are computed from word-level SMT translation scores.

\begin{table*}
\small
\centering

\begin{tabular}{|ll|c|c|}
\hline
 \textbf{Setup} && \textbf{dev} & \textbf{test} \\ \hline
 \multicolumn{2}{|l|}{SMT baseline~\cite[Travatar]{smt-travatar}}  & 19.5 & 22.2 \\  \hline
Single NMT (word) & Pure NMT & 20.3 & 22.5 \\
 & 10k-best rescoring & 22.2 & 24.5 \\
 & {\bf This work} & {\bf 22.4} & {\bf 25.2}  \\ \hline
6-Ensemble NMT (word) & Pure NMT & 22.6 & 25.0 \\
 & 10k-best rescoring & 22.4 & 25.4 \\
 & {\bf This work} & {\bf 23.9} & {\bf 26.5}  \\ \hline
Single NMT (BPE) & Pure NMT & 20.8 & 23.5 \\
 & 10k-best rescoring & 21.9 & 24.6 \\
 & {\bf This work} & {\bf 23.0} & {\bf 25.4}  \\ 
 \hline
3-Ensemble NMT (BPE) & Pure NMT & 23.3 & 25.9 \\
 & 10k-best rescoring & 22.6 & 25.1 \\
 & {\bf This work} & {\bf 24.1} & {\bf 26.7}  \\ 
 \hline
\end{tabular}
\caption{\label{tab:results-jaen} Japanese-English cased BLEU scores calculated with Moses' \texttt{multi-bleu.pl}.$^5$}
\end{table*}

\section{Experimental Setup}

We test our approach on English-German (En-De) and Japanese-English (Ja-En). For En-De, we use the WMT {\em news-test2014} (the filtered version) as a development set, and keep {\em news-test2015} and {\em news-test2016} as test sets. For Ja-En, we use the ASPEC corpus~\cite{data-aspec} to be strictly comparable to the evaluation done in the Workshop of Asian Translation (WAT).

\footnotetext{Comparable to \url{http://matrix.statmt.org/}}

The NMT systems are as described by Stahlberg et al.~\shortcite{sgnmt} using the  Blocks and Theano frameworks~\cite{blocks,theano} with hyper-parameters as in~\cite{bahdanau} and a vocabulary size of 30k for Ja-En and 50k for En-De. We use the coverage penalty proposed by Wu et al.~\shortcite{gnmt} to improve the length and coverage of translations. Our final ensembles combine five (En-De) to six (Ja-En) independently trained NMT systems.
 
Our En-De SMT baseline is a hierarchical system based on the HiFST package\footnote{\url{http://ucam-smt.github.io/}} which produces rich output lattices. The system uses rules extracted as described by de Gispert et al.~\shortcite{hifst-grammar} and a 5-gram language model~\cite{Heafield-estimate}.

In Ja-En we use Travatar~\cite{smt-travatar}, an open-source tree-to-string system. We provide the system with Japanese trees obtained using the Ckylark parser~\cite{ckylark} and train it on high-quality alignments as recommended by Neubig and Due~\shortcite{smt-elements-tree2str}. This system, which reproduces the results of the best submission in WAT 2014~\cite{neubig14wat}, is used to create a 10k-best list of hypotheses, which we convert into determinised and minimised FSAs for our work. Our Ja-En NMT models are trained on the same 500k training samples as the Travatar baseline.

The parameter $\lambda$ is tuned by optimising the BLEU score on the validation set, and we set $\Theta_i=1$ ($i=0,\dots,4$). Using the BOBYQA algorithm~\cite{bobyqa} or lattice MERT~\cite{lmert} to optimise the $\Theta$-parameters independently did not yield improvements. The beam search implementation of the SGNMT decoder\footnote{\url{http://ucam-smt.github.io/sgnmt/html/}}~\cite{sgnmt} is used in all our experiments. We set the beam size to 20 for En-De and 12 for Ja-En.

\section{Results}

\addtocounter{footnote}{+1}
\footnotetext{Comparable to \url{http://lotus.kuee.kyoto-u.ac.jp/WAT/evaluation/list.php?t=2}}

Our results are summarised in Tab.~\ref{tab:results-ende} and~\ref{tab:results-jaen}.\footnote{Instructions for reproducing our key results will be available upon publication at \url{http://ucam-smt.github.io/sgnmt/html/tutorial.html}} Our approach outperforms both single NMT and SMT baselines by up to 3.4 BLEU for En-De and 2.8 BLEU for Ja-En. Ensembling yields further gains across all test sets both for the NMT baselines and our MBR-based hybrid systems. We see substantial gains from our MBR-based method over lattice rescoring for both single and ensembled NMT on all test sets and language pairs except En-De {\em news-test2016}. On Ja-En, we report 26.7 BLEU$^5$, second to only one system (as of February 2017) that uses a number of techniques such as minimum risk training and a much larger vocabulary size which could also be used in our framework.

Our word-level NMT baselines suffer from their limited vocabulary since we do not apply post-processing techniques like UNK-replace~\cite{nmt-unkreplace}. Therefore, NMT with subword units (BPE) consistently outperforms them by a large margin. Lattice rescoring and MBR yield large gains for both BPE-based and word-based NMT. However, the performance difference between BPE- and word-level NMT diminishes with lattice rescoring and MBR decoding: rescoring with NMT often performs on the same level for both words and subword units, and MBR-based NMT is often even better with a word-level NMT baseline. This indicates that subword units are often not necessary when the hybrid system has access to a large word-level vocabulary like the SMT vocabulary.

Note that the BPE lattice rescoring system is constrained to produce words in the output vocabulary of the syntactic SMT system and is prevented from inventing new target language words out of combinations of subword units. MBR imposes a soft version of such a constraint by biasing the BPE-based system towards words in the SMT search space.

The hypotheses produced by our MBR-based method often differ from the translations in the baseline systems. For example, 77.8\% of the translations from our best MBR-based system on Ja-En cannot be found in the SMT 10k-best list, and 78.0\% do not match the translation from the pure NMT 6-ensemble.\footnote{Up to NMT OOVs.} This suggests that our MBR decoder is able to produce entirely new hypotheses, and that our method has a profound effect on the translations which goes beyond rescoring the SMT search space or fixing UNKs in the NMT output.

\begin{figure}[!t]
\centering
\includegraphics[width=1\linewidth]{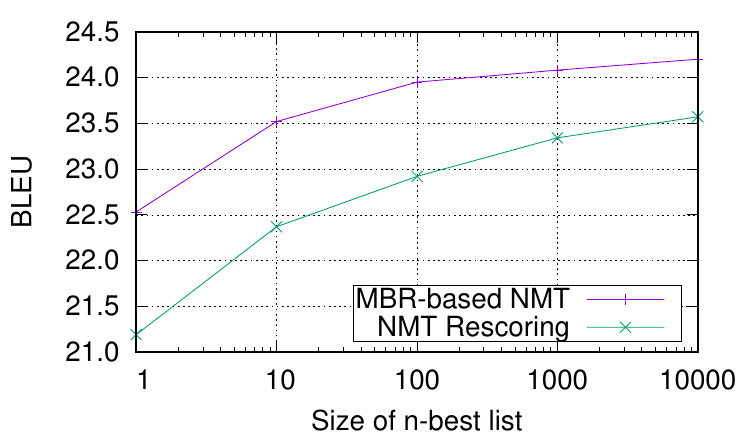}
\caption{Performance over $n$-best list size on English-German {\em news-test2015}.}
\label{fig:lat-size}
\end{figure}

Tab.~\ref{tab:results-ende} also shows that rescoring is sensitive to the size of the $n$-best list or lattice: rescoring the entire lattice instead of a 100-best list often yields a gain of 1 full BLEU point. In order to test our MBR-based method on small lattices, we compiled $n$-best lists of varying sizes to lattices and extracted $n$-gram posteriors from the reduced lattices. Fig.~\ref{fig:lat-size} shows that the $n$-best list size has an impact on both methods. Rescoring a 10-best list already yields a large improvement of 1.2 BLEU. However, the hypotheses are still close to the SMT baseline. The MBR-based approach can make better use of small $n$-best lists as it does not suffer this restriction. MBR-based combination on a 10-best list performs on about the same level as rescoring a 10,000-best list which demonstrates a practical advantage of MBR over rescoring.

\section{Related Work}

Combining the advantages of NMT and traditional SMT has received some attention in current research. A recent line of research attempts to integrate SMT-style translation tables into the NMT system~\cite{nmt+smt-dict1,nmt+smt-dict2,nmt+smt-dict3}. Wang et al.~\shortcite{nmt+smt-train} interpolated NMT posteriors with word recommendations from SMT and jointly trained NMT together with a gating function which assigns the weight between SMT and NMT scores dynamically. Neubig et al.~\shortcite{neubigAtwat2015} rescored $n$-best lists from a syntax-based SMT system with NMT. Stahlberg et al.~\shortcite{sgnmt} restricted the NMT search space to a Hiero lattice and reported improvements over $n$-best list rescoring. Stahlberg et al.~\shortcite{editdistance} combined Hiero and NMT via a loose coupling scheme based on composition of finite state transducers and translation lattices which takes the edit distance between translations into account. Our approach is similar to the latter one since it allows to divert from SMT and generate translations without derivations in the SMT system. This ability is crucial for NMT ensembles because SMT lattices are often too narrow for the NMT decoder~\cite{editdistance}. However, the method proposed by Stahlberg et al.~\shortcite{editdistance} insists on a monotone alignment between SMT and NMT translations to calculate the edit distance. This can be computationally expensive and not appropriate for MT where word reorderings are common. The MBR decoding described here does not have this shortcoming.

\section{Conclusion}

This paper discussed a novel method for blending NMT with traditional SMT by biasing NMT scores towards translations with low Bayes-risk with respect to the SMT lattice. We reported significant improvements of the new method over lattice rescoring on Japanese-English and English-German and showed that it can make good use even of very small lattices and $n$-best lists.

In this work, we calculated the Bayes-risk over non-neural SMT lattices. In the future, we are planning to introduce neural models to the risk estimation while keeping the computational complexity under control, e.g.\ by using neural $n$-gram language models~\cite{nlm,nlm-nplm} or approximations of NMT scores~\cite{rnnlm-fst,rnnlm-clustering} for $n$-gram posterior calculation.

\section*{Acknowledgments}

This work  was  supported  by the U.K.\ Engineering and Physical Sciences Research Council (EPSRC grant EP/L027623/1).

We thank Graham Neubig for providing pre-trained parsing and alignment models, as well as scripts, to allow perfect reproduction of the NAIST WAT 2014 submission. 

\bibliography{refs}
\bibliographystyle{eacl2017}

\end{document}